\def\year{2022}\relax
\documentclass[letterpaper]{article} 
\usepackage{aaai22}  
\usepackage{times}  
\usepackage{helvet}  
\usepackage{courier}  
\usepackage[hyphens]{url}  
\usepackage{graphicx} 
\def\UrlFont{\rm}  
\usepackage{natbib}  
\usepackage{caption} 
\DeclareCaptionStyle{ruled}{labelfont=normalfont,labelsep=colon,strut=off} 
\usepackage{algorithm}
\usepackage{algorithmic}

\usepackage{newfloat}
\usepackage{listings}
\floatstyle{ruled}
\newfloat{listing}{tb}{lst}{}
\floatname{listing}{Listing}
\usepackage{times}
\usepackage{latexsym}
\usepackage{booktabs} 
\usepackage{multirow} 
\usepackage{verbatimbox} 
\usepackage[T1]{fontenc}
\usepackage[utf8]{inputenc}

\usepackage{microtype}
\usepackage{graphicx}
\usepackage{enumitem}
\usepackage[textsize=scriptsize]{todonotes}
\usepackage{color}

\usepackage{pifont}
\newcommand{\cmark}{\ding{51}}%
\newcommand{\xmark}{\ding{55}}%
\newcommand{\dataset}{\texttt{TempQA-WD}}
\newcommand{\system}{SYGMA}

\title{SYGMA: System for Generalizable Modular Question Answering Over Knowledge Bases}
\author{
Sumit Neelam, Udit Sharma, Hima Karanam, Shajith Ikbal, Pavan Kapanipathi\\
Ibrahim Abdelaziz, Nandana Mihindukulasooriya, Young-Suk Lee, Santosh Srivastava \\
Cezar Pendus, Saswati Dana, Dinesh Garg, Achille Fokoue, G P Shrivatsa Bhargav\\
Dinesh Khandelwal, Srinivas Ravishankar, Sairam Gurajada, Maria Chang, \\
Rosario Uceda-Sosa, Salim Roukos, Alexander Gray, Guilherme Lima\\
Ryan Riegel, Francois Luus, L Venkata Subramaniam  \\
    IBM Research
}

\begin{document}

\maketitle

\begin{abstract}
 
Knowledge Base Question Answering (KBQA) tasks that involve complex reasoning are emerging as an important research direction. However, most KBQA systems struggle with generalizability, particularly on two dimensions: (a) across multiple reasoning types where both datasets and systems have primarily focused on multi-hop reasoning, and (b) across multiple knowledge bases, where KBQA approaches are specifically tuned to a single knowledge base. In this paper, we present \system, a modular approach facilitating generalizability  across multiple knowledge bases and multiple reasoning types. Specifically, \system~contains three high level modules: 1) KB-agnostic question understanding module that is common across KBs 2) Rules to support additional reasoning types and 3) KB-specific question mapping and answering module to address the KB-specific aspects of the answer extraction. We demonstrate effectiveness of our system by evaluating on datasets belonging to two distinct knowledge bases, DBpedia and Wikidata. In addition, to demonstrate extensibility to additional reasoning types we evaluate on multi-hop reasoning datasets and a new Temporal KBQA benchmark dataset on Wikidata, named \dataset\footnote{https://github.com/IBM/tempqa-wd}, introduced in this paper. We show that our generalizable approach has better competetive performance on multiple datasets on DBpedia and Wikidata that requires both multi-hop and temporal reasoning. 

\end{abstract}
\section{Introduction}
\label{sec:intro}
The goal of Knowledge Base Question Answering (KBQA) systems is to answer natural language questions by retrieving and reasoning over facts in a Knowledge Base (KB). KBQA has gained significant popularity both due to its practical real-world applications and challenging research problems associated with it~\cite{fu2020survey}. However, most existing datasets and research in this area are primarily focused on \textit{single/multi-hop} reasoning on a \textit{single} knowledge base~\cite{trivedi2017lc,QALD_2017,yih-etal-2016-value}. 
Consequently, this has encouraged research in methodologies that are tuned to a restricted set of reasoning types on a single knowledge base, in turn lacking generalizability~\cite{pavan2020,zou2014natural,webqsp}. In this work, we propose a modular approach that is generalizable across knowledge bases and reasoning types.   

\begin{figure}[th]
  \centering
  \includegraphics[height=4.7cm,width=0.75\linewidth]{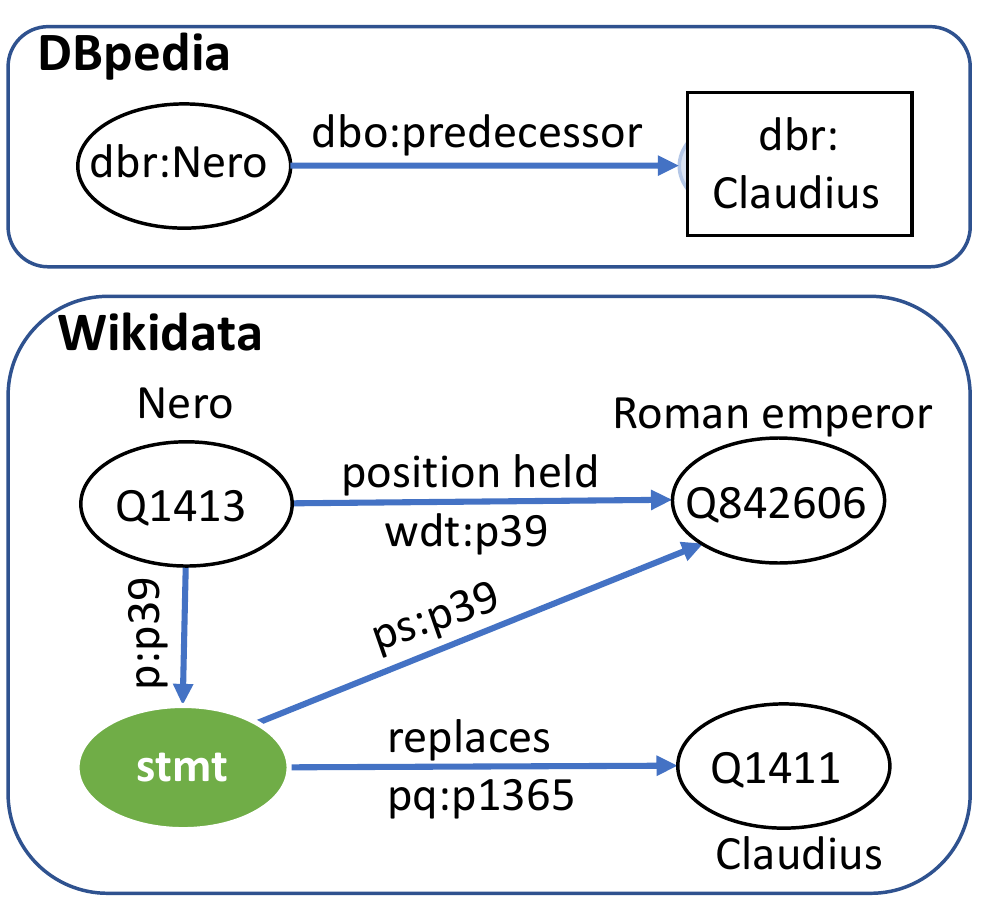}
  \caption{A description of how information about Nero's predecessor is represented in DBpedia vs. Wikidata}
  \label{fig:query_graphs_comparison} 
\end{figure}

 Open Knowledge Bases such as DBpedia~\cite{auer2007dbpedia}, Wikidata~\cite{vrandevcic2014wikidata}, and Freebase~\cite{bollacker2008freebase} form the basis of many KBQA datasets. Each of these knowledge Bases, however, have different representations. For instance: information associated to the question \textit{``Who was roman emperor before Nero?"} from DBpedia and Wikidata are shown in Figure~\ref{fig:query_graphs_comparison}. We can see that Wikidata represents properties of facts such as replaces, temporal and spatial with reification.\footnote{Information about facts are explicitly mapped by making facts as primitive nodes.} On the other hand, DBpedia manages to represent it as a simple fact with the relationship to existing entity nodes. Handling such different representations is an unexplored challenge and requires to be addressed for a generalizable KBQA approach.
  The second aspect where KBQA approaches fail to generalize is in handling complex reasoning types such as temporal and spatial reasoning. Table~\ref{tab:multi-hop} shows examples of questions that require different types of reasoning to answer them. Most research have focused on multi-hop reasoning questions~\cite{bordes2015large,lcquad2,berant2013}, lacking both approaches and datasets that are adaptable to complex reasoning types. Particularly a single approach that can answer questions with different reasoning types.

In this paper, we present a modular approach called \textbf{\system}~(\textbf{Sy}stem for \textbf{G}eneralizable and \textbf{M}odular question \textbf{A}nswering over knowledge bases), that is built on a framework adaptable to different KB representations and reasoning types. \system~ introduces intermediate representations based on lambda calculus to adapt to new reasoning types and representations. \system~, following the recent advances in KBQA such as NSQA~\cite{pavan2020} and ReTraCk~\cite{chen-etal-2021-retrack}, is a modular approach with independent Abstract Meaning Representation, Entity Linking, and Relation Linking modules. In contrast to NSQA which is tuned specifically to DBpedia-based KBQA datasets, \system~is generalizable to multiple knowledge bases and multiple reasoning types. In order to evaluate our generalizable approach, we consider datasets that have: (a) DBpedia and Wikidata as knowledge bases, and (b) multi-hop and temporal as the reasoning types. 

In pursuit of this goal, we also address the lack of temporal reasoning datasets on knowledge bases that we are considering in this work, specifically Wikidata. We annotate and create \dataset; a Wikidata-based dataset focused on temporal reasoning. \dataset~is a parallel dataset to TempQuestions dataset~\cite{jia2018a} which is a Freebase based temporal reasoning dataset derived from three other datasets \textit{Free917} \cite{cai2013}, \textit{WebQuestions} \cite{berant2013} and \textit{ComplexQuestions} \cite{bao2016}. While the Freebase dataset consists of question answer pairs along with temporal signal classifications, our dataset is annotated with intermediate SPARQL queries enabling evaluation of different modules in a KBQA pipeline. 


\begin{table}[htb]
\begin{small}
\centering
\begin{tabular}{l|l}
\toprule
\begin{tabular}[l]{@{}l@{}}
    \textbf{Category}\\
\end{tabular} 
& 
\textbf{Example} \\
\midrule

\begin{tabular}[l]{@{}l@{}}
    Single-Hop
\end{tabular} 
& 

\begin{tabular}[c]{@{}l@{}}

\emph{ Who directed Titanic Movie?}\\
{\bf SPARQL:}
select distinct ?a where \{\\
  wd:Q44578 wdt:P57 ?a\}\\

    \end{tabular} \\
\hline
\begin{tabular}[l]{@{}l@{}}
    Multi-hop\\
\end{tabular} 
& 

\begin{tabular}[c]{@{}l@{}}

\emph{ Which movie is directed by James Cameron }\\
\emph{ starring Leonardo DiCaprio?  }\\
{\bf SPARQL:}
select distinct ?a where \\
  \{?a wdt:P57 wd:Q42574. \\
  ?a wdt:P161 wd:Q38111. \} \\
   \end{tabular} \\
\hline
\begin{tabular}[l]{@{}l@{}}
    Temporal
\end{tabular} 
& 
\begin{tabular}[c]{@{}l@{}}
\emph{ Who was the US President }\\
\emph{ during cold war?  }{\bf SPARQL:}
in Figure~\ref{fig:example} \\
\end{tabular} \\
\bottomrule
\end{tabular}
\caption{Examples of Single-hop, Multi-hop and Temporal reasoning questions on Wikidata.}
\label{tab:multi-hop}
\end{small}
\end{table}

In summary, the main contributions of this paper are:

\begin{itemize}
      \item A modular, generalizable approach, called \system, for KBQA with lambda calculus based intermediate representations that enable adaptation to: (a) multiple knowledge bases, specifically DBpedia and Wikidata, and (b) multiple reasoning types such as multi-hop and temporal reasoning. 
     \item A benchmark dataset, called \dataset, for building temporal-aware KBQA systems on \textit{Wikidata} with parallel annotations on \textit{Freebase}. 
    \item Experimental results show \system~ achieves state-of-the-art numbers on LC-QuAD 1.0, WebQSP-WD, and comparable numbers on QALD-9 dataset. We also present baseline numbers for Simple WebQuestions and new dataset introduced in this paper; \dataset.
 \end{itemize}

\section{Related Work}
\begin{table*}
\centering
\resizebox{2\columnwidth}{!}{%
\begin{tabular}{l|l|l|c|c|c}
\toprule
\textbf{Datasets} & \textbf{Knowledge Base} & \textbf{Emphasis} & \textbf{SPARQL} & \textbf{Intermediate} & \textbf{Natural questions}\\
 &  &  & \textbf{} &  \textbf{Evaluation} & \\ 
\midrule  
QALD-9 & DBpedia & Multi-hop & \cmark & \cmark & \cmark \\
LC-QuAD 1.0 & DBpedia & Multi-hop & \cmark & \cmark & \xmark \\
LC-QuAD 2 & DBpedia, Wikidata & Multi-hop & \cmark & \cmark & \xmark \\
Simple Questions & Freebase, DBpedia, Wikidata & Single-hop & \cmark & \cmark & \cmark \\
Web Questions & Freebase
& Multi-hop
& \xmark
& \xmark
& \cmark \\
WebQSP
& Freebase, Wikidata
& Multi-hop
& \cmark
& \cmark
& \cmark \\
Complex Web Questions
& Freebase
& Multi-hop
& \xmark
& \xmark
& \cmark \\
TempQuestions
& Freebase
& Temporal
& \xmark
& \xmark
& \cmark \\
\midrule
\midrule
\textbf{\dataset}
& \textbf{Freebase, Wikidata}
& \textbf{Temporal}
& \cmark
& \cmark
& \cmark \\
\bottomrule
\end{tabular}
}
\caption{This table compares most of the KBQA datasets based on features relevant to the work presented in this paper. 
}
\label{tab:related_work_datasets}
\end{table*}

{\bf KBQA Systems:} 
KBQA approaches can be primarily categorized into two groups: (a) Question-to-entities: where techniques output the answer entities from the knowledge graph ignoring the SPARQL query~\cite{saxena-etal-2020-improving,sun-etal-2018-open,qamp}, and (b) semantic parsing based: where approaches output intermediate SPARQL queries (or logic forms) that can retrieve the answer from the knowledge base~\cite{singh2018frankenstein,pavan2020,gAnswer}. First, in question-to-entities category,~\cite{qamp} leverages a message passing technique on a two hop graph from the entities mentioned in the question to retrieve answers from the knowledge graph. 
~\cite{saxena-etal-2020-improving} encodes both text and entities, text based on language models and entities based on knowledge graph embeddings~\cite{complexE} and shows that text can help KBQA in an incomplete setting. In contrast to question-to-entities approach, the semantic parsing based approaches improves interpretability and facilitates evaluation of different modules as shown by Frankestein~\cite{singh2018frankenstein}, and NSQA~\cite{pavan2020}. Both, Frankestein~\cite{singh2018frankenstein}, and NSQA~\cite{pavan2020} follow a pipeline based approach with the differentiating factor being the use of Abstract Meaning Representation as a starting point by NSQA. We build on top of these representational efforts and introduce $\lambda$-expressions with the required additional functions for adapting to new knowledge graphs and new reasoning types. 




{\bf KBQA Datasets:} Over the years,  many  question answering datasets have been developed for KBQA, such as Free917~\cite{cai2013}, SimpleQuestions~\cite{journals/corr/BordesUCW15}, WebQuestions~\cite{berant2013}, QALD-9~\cite{QALD_2017}, LC-QuAD 1.0~\cite{trivedi2017lc}, and LC-QuAD 2.0~\cite{lcquad2}. In Table~\ref{tab:related_work_datasets}, we compare each of these datasets across the following features: (a) underlying KB, including subsequent extensions, e.g. Wikidata~\cite{wikidata-benchmark} and DBpedia~\cite{azmy-etal-2018-farewell} based versions of SimpleQuestions, as well as the Wikidata subset of WebQSP ~\cite{webqsp}; 
(b) reasoning types that are emphasized in the dataset;
(c) availability of SPARQL queries, entities, and relationships for intermediate evaluation; and (d) the use of question templates, which can often generate noisy, unnatural questions.
As Table~\ref{tab:related_work_datasets} shows, our dataset is distinguished from prior work in its emphasis on temporal reasoning, its application to both Freebase and Wikidata, and its annotation of intermediate representations and SPARQL queries. The most relevant KBQA dataset to our work is TempQuestions~\cite{jia2018a}, upon which we base \dataset, as  described in dataset section. CRONQUESTIONS~\cite{DBLP:conf/acl/SaxenaCT20} is another temporal QA dataset proposed recently over a small subset of Wikidata. It has template based questions targeted at temporal KG embeddings and QA.

\section{~\system: System Description}
\label{sec:baseline}
\begin{figure*}[t!]
  \centering
  \includegraphics[width=0.9\linewidth]{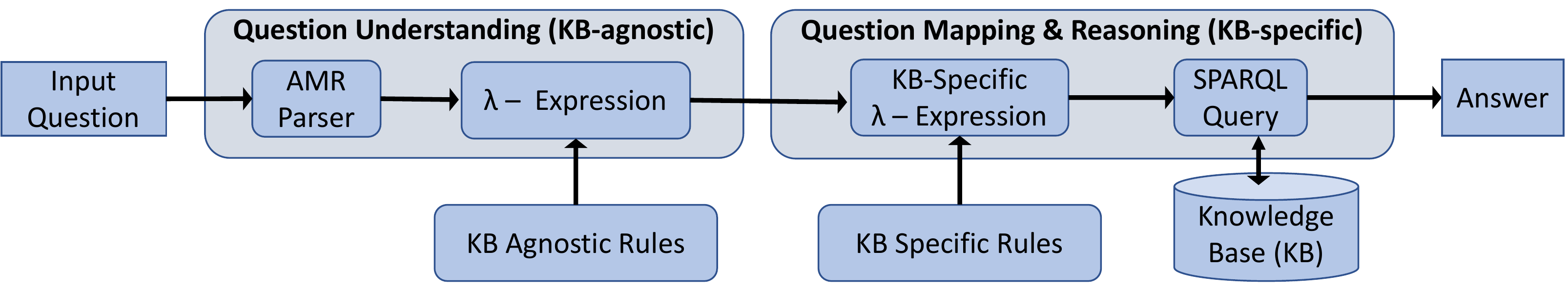}
  \caption{Architecture of ~\system\ that shows the pipeline with independent modules and intermediate representations. The intermediate representations are supported by appropriate heuristics to facilitate generalizability.}
  \label{fig:pipeline} 
\end{figure*}

\begin{figure*}[t!]
  \centering
  \includegraphics[width=0.9\linewidth]{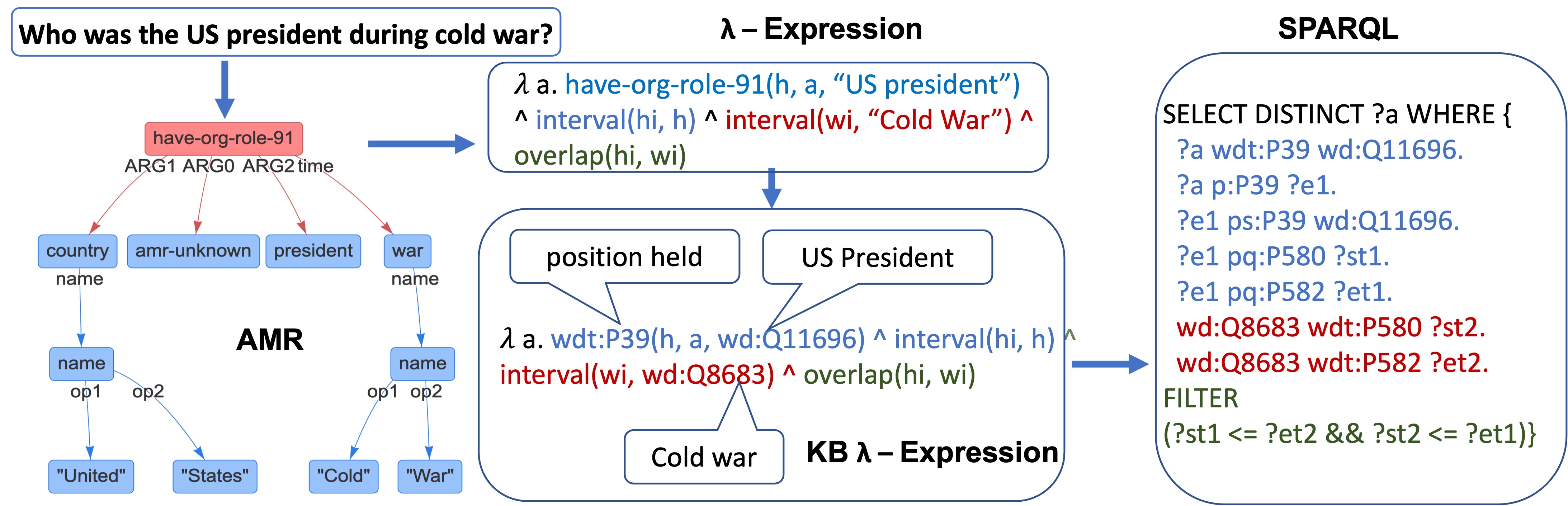}
  \caption{An illustration of the outputs at the intermediate stages of the pipeline in~\system}  
  \label{fig:example} 
\end{figure*}

Motivated by \citet{pavan2020},~\system~is designed as a neuro-symbolic system, avoiding the need for end-to-end training. Figure \ref{fig:pipeline} shows the overall KBQA pipeline with independent modules and their intermediate representations. The approach is designed with the goal of generalization across multiple KBs and reasoning types. Supporting generalization is accomplished in two stages as shown in Figure~\ref{fig:pipeline},  driven by an example in Figure~\ref{fig:example}. The first is a \textit{KB-agnostic question understanding} stage, which takes in a natural language question as input and outputs an intermediate  representation that can be common across different KBs. The second is \textit{question mapping and reasoning} stage where, first, the necessary heuristics to introduce templates for new reasoning types are applied. Next, the KB-agnostic representation is mapped to the vocabulary of the KG to output KB-specific representation that has a deterministic mapping to SPARQL. We use Wikidata and DBpedia to show generalization across KBs, and use multi-hop and temporal reasoning to evaluate across reasoning types.  
Before we get into the details of these modules and intermediate representations, we give a brief description of the Lambda calculus and the temporal functions used by \system\ to generate its intermediate logic representation.

\noindent
\textbf{Lambda Calculus}
$\lambda$-calculus, by definition, is considered the smallest universal programming language that expresses any computable function.
In particular, we have adopted \textit{Typed} $\lambda$-\textit{Calculus} presented in \cite{zettlemoyer2012learning} which support addition of new higher order functions necessary for handling various reasoning types. 
We use constants and logical connectives like AND, OR, NEGATION and functions like argmin, argmax, count, etc., presented in this work. Apart from this, we also added new temporal functions to demonstrate the system's adaptability to support new reasoning types.
For example, consider the following question and its corresponding logical form:  
\begin{center}
\begin{tabular}{l c}
 Question: & When was Barack Obama born?\\ 
 Logical Form: & \begin{tabular}{l l}$\lambda$\textit{t. born(b,``Barack Obama")} $\wedge$\\\textit{interval(t, b)} \end{tabular}\\  
\end{tabular}
\end{center}
Here, \textit{b} is instance variable for event \textit{born(b, ``Barack Obama")} and \textit{interval(t, b)} finds time for the event denoted by \textit{b}. Variable \textit{t} is unknown which is marked as $\lambda$ variable.

\noindent
\textbf{Temporal Functions:} We introduce \textit{interval, overlap, before, after, teenager, year}; where \textit{interval} gets time interval associated with event and \textit{overlap, before, after} are used to compare temporal events. \textit{teenager} gets teenager age interval for a person, and \textit{year} return year of a date.\footnote{Although there are many other possible temporal functions for specific English words, such as \textit{adult} and \textit{tween}, we defined \textit{teenager} because it was featured in the datasets explored here.}


\subsection{KB-Agnostic Question Understanding}
The modules in this stage aim at deriving logical expressions of the natural language question that is common for all the knowledge bases.
In particular, we use the formalism of $\lambda$-calculus for logical representation, i.e., to map questions to their corresponding $\lambda$-expressions representing the semantics.   

\subsubsection{AMR}
In order to have a generic parse (logical expression) that is common across KBs,~\system~performs initial language understanding using Abstract Meaning Representation (AMR) \cite{banarescu2013}, a semantic representation language based on PropBank. AMR  encodes the meaning of a sentence into a rooted directed acyclic graph where nodes represent concepts and edges represent relations. 
We adopt an action pointer transformer architecture of \citet{JoeZhou2021} for transition-based AMR parsing and self-training technique of \citet{YoungSuk2020} for domain adaptation to KBQA. 

AMR provides generic  representation that can be used for multi-hop reasoning~\cite{pavan2020}. However for temporal expressions with \textit{:time} relation, we have to augment the AMR annotations with implicit predicates that cannot be captured by ellipsis and/or re-entrancies. 
An example of AMR annotation for the question \textit{Who was US president during cold war?} is given in Figure~\ref{fig:example}. The representation encodes time edge with cold war event as sub-event of time. Also in case of before/after constraints it explicitly captures the constraint as part of the time edge.
If there are no explicit constraints like before/after or ordinal in the time edge, we treat them as overlap constraints. 
\subsubsection{KB-Agnostic Lambda Expression}
AMR is a fairly granular representation. However, for a KBQA system, specifically on KGs such as DBpedia, Wikidata, and Freebase, such granularity can add noise. Therefore to construct KB-agnostic  $\lambda$-expressions of the questions from their corresponding AMRs and their identified entity and relation mentions, we use the transformational heuristics described in  Table~\ref{tab:amrTranslation}. 
The table shows the transformation from AMR frame (high level) to the corresponding Lambda expression. The rule type describes where the rule is in general applied to all reasoning (base) vs temporal. 
We handle these as conjunction of multiple triples in the KB with some projection variable and/or numerical operation to get the final answer. We use AMR unknown/imperative constructs to identify the projection variables and AMR polarity/interrogative frames for queries which fall into boolean answers or ASK questions in SPARQL. 

The next set of rules present in Table~\ref{tab:amrTranslation} are few sample template rules used to capture temporal events and constraints to help in temporal reasoning.  Figure~\ref{fig:example} gives an example question that falls into the temporal overlap rule in the table and how the AMR constructs are used to split the events is highlighted with nodes being bold in the table. Complete set of rules used can be found in the supplementary material. Note that we covered temporal reasoning as an additional form of reasoning in the current system. However,  more reasoning templates like spatial reasoning with additional operators like \textit{coordinates(), south()} ...etc can be added. Table~\ref{tab:amrTranslation} shows an example template for spatial questions such as \textit{"which states are to the south of California?"}.

\begin{table*}[htb] \setlength\tabcolsep{7pt}
    \centering
    \begin{small}
    \resizebox{\textwidth}{!}{
    \begin{tabular}{lll}
    
        \toprule
        \begin{tabular}[l]{@{}l@{}}
           \textbf{Rule}\\
           \textbf{Type}
        \end{tabular} 
        &
        \textbf{AMR A = (v/frame …)}
        &
        \textbf{Lambda Expression L = $\psi$(v)} 
        \\ 
        \midrule

        \begin{tabular}[l]{@{}l@{}}
           Base
        \end{tabular} 
         &
         \begin{tabular}[l]{@{}l@{}}
            (\textbf{v}/frame :arg0(\textbf{v0}/frame0) … :argn(\textbf{vn}/framen))
        \end{tabular} 
         &
         \begin{tabular}[c]{@{}l@{}}
            frame(\textbf{v}, \textbf{v0}, … \textbf{vn}) $\land$ $\psi$(\textbf{v0}) $\land$ … $\psi$(\textbf{vn}) 
        \end{tabular} 
        \\
         \noalign{\smallskip}

         \begin{tabular}[l]{@{}l@{}}
           Base
        \end{tabular} 
         &
         \begin{tabular}[l]{@{}l@{}}
            (\textbf{v}/frame :\textbf{arg1(a/amr-unknown)}  … :argn(vn/framen))
        \end{tabular} 
         &
         \begin{tabular}[c]{@{}l@{}}
           $\lambda$a. $\psi$(\textbf{v})
        \end{tabular} 
        \\
         \noalign{\smallskip}



         
         \begin{tabular}[l]{@{}l@{}}
           Numerical
        \end{tabular} 
         &
         \begin{tabular}[l]{@{}l@{}}
            (\textbf{v}/frame :arg0(\textbf{v0}/frame0 :\textbf{quant(a/amr-unknown})) …\\ :argn(vn/framen))
        \end{tabular} 
         &
         \begin{tabular}[c]{@{}l@{}}
            count($\lambda$\textbf{v0}. $\psi$(\textbf{v}))
        \end{tabular} 
        \\
         \noalign{\smallskip}

         \begin{tabular}[l]{@{}l@{}}
          Temporal
        \end{tabular} 
         &
         \begin{tabular}[l]{@{}l@{}}
            (\textbf{v}/frame :arg0(v0/frame0) … :argn(vn/framen)\\
            \textbf{:time(a/amr-unknown))}
        \end{tabular} 
         &
         \begin{tabular}[c]{@{}l@{}}
          $\lambda$ev. $\psi$(\textbf{v}) $\land$ interval(ev, \textbf{v}) 
        \end{tabular} 
        \\
         \noalign{\smallskip}

        \begin{tabular}[l]{@{}l@{}}
           Temporal
        \end{tabular} 
         &
         \begin{tabular}[l]{@{}l@{}}
            (\textbf{v}/frame :arg0(a/amr-unknown) … :argn(vn/framen)\\
            \textbf{:time(b/before} :op1(\textbf{n}/nested-frame)))
        \end{tabular} 
         &
         \begin{tabular}[c]{@{}l@{}}
           argmax($\lambda$a. $\psi$(\textbf{v}) , $\lambda$a. $\lambda$ev. $\psi$(\textbf{n}) $\land$ interval(ev, \textbf{v}) \\
           $\land$ interval(en, \textbf{n}) $\land$ \textbf{before}(ev, en), 0, 1)
        \end{tabular} 
        \\
         \noalign{\smallskip}


        \begin{tabular}[l]{@{}l@{}}
           Temporal
        \end{tabular} 
         &
         \begin{tabular}[l]{@{}l@{}}
            (\textbf{v}/frame :arg0(a/amr-unknown) … :argn(vn/framen)\\
            \textbf{:time}(\textbf{n}/nested-frame))
        \end{tabular} 
         &
         \begin{tabular}[c]{@{}l@{}}
           $\lambda$a. $\psi$(\textbf{v}) $\land$ $\psi$(\textbf{n}) $\land$ interval(ev, \textbf{v}) $\land$ interval(en, \textbf{n})\\
           $\land$ \textbf{overlap}(ev, en)
        \end{tabular} 
        \\
         \noalign{\smallskip}



     
         \begin{tabular}[l]{@{}l@{}}
           Spatial
        \end{tabular} 
         &
         \begin{tabular}[l]{@{}l@{}}
            (\textbf{b}/\textbf{be-located-at-91} :arg0(a/amr-unknown),\\ \textbf{:mod(s/south)} :op1(\textbf{n}/nested-frame))
        \end{tabular} 
         &
         \begin{tabular}[c]{@{}l@{}}
           $\lambda$a. $\psi$(\textbf{b})$\land$  $\psi$(\textbf{n}) $\land$ coordinate(cb, \textbf{b}) $\land$ coordinate(cn, \textbf{n})\\
           $\land$ \textbf{south}(cb, cn)
        \end{tabular} 
        \\
         \noalign{\smallskip}

    \bottomrule
    \end{tabular}
    }
    \end{small}
    \caption{AMR to Lambda Translation}
    \label{tab:amrTranslation}
\end{table*}

\subsection{Question Mapping and Reasoning}
This is the second stage which comprises of modules that transform the KB-agnostic $\lambda$-expression of the questions into KB-specific $\lambda$-expression. These modules are entity linking and relationship linking that are specific to the underlying KB along with any KB specific rules to handle special forms of transformation or reasoning. The modules are interchangeable to different knowledge graphs. Currently, we demonstrate the adaptability of these modules to DBpedia and Wikidata. 

\subsubsection{KB-Specific Lambda Expression:}
Structurally, it is similar to $\lambda$-expression with all entities and relations mapped to KB entities and relations respectively. 

\noindent\textbf{\emph{Entity Linking:}} The goal of Entity Linking is to map entity mentions as captured in the $\lambda$-expression of the question to their corresponding KB-specific entities. We use a recently proposed zero-shot entity linking approach called BLINK \cite{BLINK}. For a question where entity mentions are already identified, bi-encoder piece of the BLINK is used to predict top-K entities. For this prediction, we use an entity dictionary of 5.9M English Wikipedia entities, mapped to their corresponding Wikidata  and DBpedia entities. 
\noindent\textbf{\emph{Relation Linking:}}
\system's Relation linking component
takes in the question text and an AMR graph as input and returns a ranked list of KG relationships. For instance, given a question such as "Who was the US President during cold war?" (see Fig~\ref{fig:example}) and the corresponding AMR, the goal of the relation linking component is to find the corresponding Wikidata KB relations \textit{position held (P39)}, \textit{start time (P580)}, \textit{end time (P582)}. For this task, we use state-of-the-art AMR-based relation linking approach with models built for both DBpedia and Wikidata~\cite{naseem-etal-2021-semantics}.



\noindent\textbf{\textit{{SPARQL Query:}}}
This module maps KB-specific  $\lambda$-expressions onto SPARQL queries through a deterministic approach. Each KB-specific $\lambda$-expression construct is mapped to an equivalent SPARQL construct, as rules given in Table~\ref{tab:kbSpecificTranslation}.
Each of lambda expressions contains one or more terms such that each term $T_i$ is comprised of one or more predicates connected via $\land$ or $\lor$. Translation of different predicates is present in Table~\ref{tab:kbSpecificTranslation}. Some of the KB-spe cific aspects like handling reification in Wikidata are shown in the Table.
To get the time interval in case of reified events, \textit{Start time(wdt:P580)}, \textit{end time(wdt:P582)}, or \textit{point in time(wdt:P585)} connected to intermediate statement node are used. 

\begin{table*}[htb] \setlength\tabcolsep{7pt}
    \centering
    \begin{small}
    \resizebox{\textwidth}{!}{%
    \begin{tabular}{lll} 
        \toprule
        \textbf{Type} &\textbf{ Expression/Predicate E} & \textbf{SPARQL S = $\phi$(E)} \\ 
        \midrule
        
        \begin{tabular}[l]{@{}l@{}}
            $\lambda$ abstraction
        \end{tabular} 
         &
         \begin{tabular}[l]{@{}l@{}}
            $\lambda$x.T
        \end{tabular} 
         &
         \begin{tabular}[c]{@{}l@{}}
          SELECT DISTINCT ?x WHERE \{ $\phi$(T) \}
        \end{tabular} 
        \\
         \noalign{\smallskip}

        \begin{tabular}[l]{@{}l@{}}
          Count expression
        \end{tabular} 
         &
        \begin{tabular}[l]{@{}l@{}}
             count($\lambda$x.T)
        \end{tabular} 
         &
         \begin{tabular}[c]{@{}l@{}}
          SELECT (COUNT(?x) AS ?c) WHERE \{ $\phi$(T) 
          \}
        \end{tabular}
        \\
         \noalign{\smallskip}


        \begin{tabular}[l]{@{}l@{}}
          Argmax expression
        \end{tabular} 
         &
        \begin{tabular}[l]{@{}l@{}}
             argmax($\lambda$x.T1, $\lambda$x.$\lambda$y. T2, \textit{O}, \textit{L})
        \end{tabular} 
         &
         \begin{tabular}[c]{@{}l@{}}
          SELECT DISTINCT ?x WHERE \{
           $\phi$(T1)  $\phi$(T2) \\
          \} ORDER BY DESC(?yStart) LIMIT \textit{L} OFFSET \textit{O}
        \end{tabular}
        \\\noalign{\smallskip}


        \begin{tabular}[l]{@{}l@{}}
          KB Predicate
        \end{tabular} 
         &
        \begin{tabular}[l]{@{}l@{}}
            IRI$_p$(r, s$|$IRI$_s$, o$|$IRI$_o$)
        \end{tabular} 
         &
         \begin{tabular}[c]{@{}l@{}}
          ?s$|$IRI$_s$ IRI$_p$ ?o$|$IRI$_o$.
        \end{tabular} 
        \\
         \noalign{\smallskip}

        \begin{tabular}[l]{@{}l@{}}
            Interval predicate\\ for reified facts
        \end{tabular} 
         &
        \begin{tabular}[l]{@{}l@{}}
             wdt:PID(e, s$|$wd:QID$_1$, o$|$wd:QID$_2$)\\
             $\land$ interval(e$^i$, e)
        \end{tabular} 
         &
         \begin{tabular}[c]{@{}l@{}}
          ?s$|$wd:QID$_1$ p:PID ?e. ?e ps:PID ?o$|$wd:QID$_2$.\\
          ?e \textit{pq:P580} ?eiStart. ?e \textit{pq:P582} ?eiEnd.
        \end{tabular} 
        \\
         \noalign{\smallskip}




        \begin{tabular}[l]{@{}l@{}}
          Overlap predicate
        \end{tabular} 
         &
        \begin{tabular}[l]{@{}l@{}}
             overlap(e1$^i$ e2$^i$)
        \end{tabular} 
         &
         \begin{tabular}[c]{@{}l@{}}
          FILTER(?e1iStart<=?e2iEnd \&\&
          ?e2iStart<=?e1iEnd)
        \end{tabular} 
        \\
         \noalign{\smallskip}

        \begin{tabular}[l]{@{}l@{}}
            Before predicate
        \end{tabular} 
         &
         \begin{tabular}[l]{@{}l@{}}
             before(e1$^i$, e2$^i$)
        \end{tabular} 
         &
         \begin{tabular}[c]{@{}l@{}}
          FILTER(?e1iEnd<=?e2iStart)
        \end{tabular}
        \\
        \noalign{\smallskip}

        \begin{tabular}[l]{@{}l@{}}
             After predicate
        \end{tabular} 
         &
        \begin{tabular}[l]{@{}l@{}}
             after(e1$^i$, e2$^i$)
        \end{tabular} 
         &
         \begin{tabular}[c]{@{}l@{}}
          FILTER(?e1iStart>=?e2iEnd)
        \end{tabular}
        \\
        \noalign{\smallskip}
    \bottomrule
    \end{tabular}
    }
    \end{small}
    \caption{Translation of KB Specific Lambda Expression}
    \label{tab:kbSpecificTranslation}
\end{table*}


    \section{TempQA-WD Dataset}
\label{sec:dataset}
In order to evaluate our approach on temporal reasoning, specifically on DBpedia and Wikidata as KGs, we require a dataset to be based on one of these two KGs. \textit{TempQuestions} \cite{jia2018a} is the first KBQA dataset intended to focus specifically on temporal reasoning. However, TempQuestions is based on Freebase.
 We adapt \textit{TempQuestions} to Wikidata\footnote{Creating a parallel dataset on two KGs (DBpedia and Wikidata) is non trivial and labor intensive. Therefore, we opted to create a dataset based on Wikidata alone since it is the most up-to-date KG.} to create a temporal QA dataset that has three desirable properties. First, we create a generalizable benchmark that has parallel answer annotations on two KBs. Second, we take advantage of Wikidata's evolving, up-to-date knowledge. Lastly, we enhance \textit{TempQuestions} with SPARQL queries which was missing in original dataset. We also add entity, relation, intermediate lambda expression annotations for a subset of the dataset that are used by the~\system.

 


There has been two previous attempts at transferring Freebase-QA questions to \textit{Wikidata}; namely WebQSP-WD~\cite{webqsp} and SimpleQuestions-WD(SWQ-WD)~\cite{diefenbach2017wdaqua}. 
SWQ-WD contains only single triple questions whereas WebQSP-WD have only the final question answers that map directly to corresponding entities in Wikidata. However, as stated ~\cite{webqsp}, one challenge is that not all \textit{Freebase} answers can be directly mapped to entities in \textit{Wikidata}. For example, the Freebase answer annotation for the question ``When did Moscow burn?''  is ``1812 Fire of Moscow'', despite the year being entangled with the event itself. In contrast, Wikidata explicitly represents the year of this event, with an entity for ``Fire in Moscow'' and an associated year of ``1812''. Thus, a direct mapping between the two answers is not possible, as it would amount to a false equivalence between ``1812 Fire of Moscow'' and ``1812''.

In order to address such issues, 
we enlisted a team of annotators to manually create and verify SPARQL queries, 
ensuring not only that the SPARQL formulation was correct, but that the answers accurately reflected the required answer type (as in the ``Fire in Moscow'' example above) and the evolving knowledge in Wikidata.
Having SPARQL queries also facilitates intermediate evaluation of the approaches that use semantic parsing to directly generate the query or the query graph, 
increasing interpretability and performance in some cases~\cite{webqsp}. 

\noindent
\textbf{Dataset Details}
\begin{table}[ht]
\begin{small}
\centering
\begin{tabular}{c|c|c|c}
\toprule
Dataset & Size & Answer & Add'l Details \\
\midrule
\textit{TempQuestions(Freebase)} & 1271 & \textit{Freebase} & A \\
\midrule
\textit{TempQA-WD (Wikidata)} & 839 & \textit{Wikidata} & A + B \\
\midrule
\textit{A Subset of TempQA-WD} & 175 & \textit{Wikidata} & A + B + C \\
\bottomrule
\end{tabular}
\caption{Benchmark dataset details. \textit{TempQuestions} denote original dataset with \textit{Freebase} answers \cite{jia2018a}. Second row is the subset adapted to \textit{Wikidata} and the third row is the \textit{devset} taken out of it.
Set-A denote \textit{TempQuestions} with- \{temporal signal, question type, data source\}. Set-B -\{Wikidata SPARQL , answer, category\}. Set-C-\{AMR, $\lambda$-expression, entities,  relations, KB-specific $\lambda$-expression\} }
\label{tab:benchmark}
\end{small}
\end{table}
Table~\ref{tab:benchmark} gives details of our new benchmark dataset. We took all the questions from \textit{TempQuestions} dataset (of size $1271$) and chose a subset for which we could find \textit{Wikidata} answers. This subset has 839 questions that constitute our new dataset, \dataset. We annotated this set with their corresponding \textit{Wikidata} SPARQL queries and \textit{Wikidata} answers. We also retained the \textit{Freebase} answers from the \textit{TempQuestions} dataset effectively creating  parallel answers from two KBs. Additionally, we  added a question complexity label to each question, according to the reasoning complexity required to answer the question. Details of categorization are present in supplementary material. Within this dataset, we chose a subset  of $175$ questions for detailed annotations as described in Set-C in the table description leaving $664$ question as test set. 

\section{Evaluation}
\label{sec:eval}
\begin{table*}[tb]
\centering
\small
\begin{tabular}{l|ccc|ccc|ccc|ccc|ccc}
\toprule
KB~~~~~~~~~~~~~$\rightarrow$ & \multicolumn{6}{|c|}{DBpedia} & \multicolumn{9}{|c}{Wikidata} \\ 
\midrule
DataSet~~~~~~$\rightarrow$& \multicolumn{3}{|c|}{LC-QuAD 1.0} &\multicolumn{3}{|c|}{QALD-9} &\multicolumn{3}{|c|}{WebQSP-WD} & \multicolumn{3}{|c|}{SWQ-WD} &\multicolumn{3}{|c}{ TempQA-WD}\\
 Reasoning~~$\rightarrow$& \multicolumn{3}{|c|}{multi-hop} &\multicolumn{3}{|c|}{multi-hop} &\multicolumn{3}{|c|}{multi-hop} & \multicolumn{3}{|c|}{single-hop} &\multicolumn{3}{|c}{ Temporal}\\
\hline
System~~$\downarrow$ & P & R & F1 & P & R & F1& P & R & F1& P & R & F1& P & R & F1 \\
\hline
WDAqua   &  0.22 & 0.38  &  0.28  & 0.26 & 0.26 & 0.25 & - & - & - & - & - & - & - & - & - \\
gAnswer   & -  & - & - &  0.29 &   0.32   & 0.29 & - & - & - & - & - &- & - & - & - \\
QAMP   & 0.25 & 0.50 &  0.33  & - & - & - & - & - & - & - & - &- & - & - & - \\
NSQA   & 0.44 &  0.45   & 0.44  & \bf 0.31 & \bf 0.32 & \bf 0.31&  - & - & -  & - & - &- & - & -  & - \\
GGNN   & - &  -   & -  & - & - & - &   0.27 & 0.32 & 0.26  & - & - &- & - & -  & - \\
\system  & \textbf{0.47} &\textbf{0.48}  & \textbf{0.47} & 0.29 & 0.30 & 0.29  & \textbf{0.32}& \textbf{0.36}& \textbf{0.31} &0.42 &0.55 &0.44 & 0.32 & 0.34 & 0.32  \\
\bottomrule
\end{tabular}
\caption{ \system's Performance across knowledge bases and reasoning types. Baseline numbers are taken from \citet{pavan2020} and \citet{C18-1280}. P-Precision, R-Recall}
\label{table:results}
\end{table*}
\subsubsection{Implementation Details:}
We implemented our pipeline using a Flow Compiler
~\cite{chakravarti-etal-2019-cfo} stitching individual modules exposed as gRPC services. We defined the ANTLR grammar to define $\lambda$-expressions.
KB-Specific $\lambda$-expression to SPARQL module is implemented in Java using Apache Jena
~SPARQL modules, to create SPARQL objects and generate the final SPARQL query that is run on target KB end point. 
The rest of the modules are implemented in Python and are exposed as gRPC services.  

\subsubsection{Datasets:}
\label{sec:data}
We considered two different KBs namely DBpedia and Wikidata to evaluate our system. We did not consider Freebase as its no longer actively maintained and is not up-to-date. 
Along with the new benchmark dataset \dataset~introduced in this paper, we also evaluate our baseline system on two other benchmark datasets adopted to Wikidata, to evaluate its effectiveness beyond temporal questions. They are SWQ-WD
~\cite{wikidata-benchmark}, which consists of 14894 train and 5622 test set questions, and WebQSP-WD
~\cite{C18-1280}, which consists of 2880 train and 1033 test set questions.
For DBpedia, we used QALD-9~\cite{QALD_2017}, that has 408 training and 150 test questions, and LC-QuAD 1.0~\cite{trivedi2017lc} data set, that has 4k training set and 1k test set. Each of these datasets operate on their own version of the DBpedia and SPARQLs provided and we used the same instance of DBpedia for evaluation.
The baseline system is tuned with dev sets of 175 from TempQA-WD and 200 from SWQ-WD train and 200 from LC-QuAD 1.0 and evaluated on all the the five test sets. 


\subsubsection{Baselines}
We Compare ~\system~ with various KBQA baseline systems as given in Table~\ref{table:results}. To our knowledge there is no other system that works across both DBpedia and Wikidata. We took the numbers reported from NSQA~\cite{pavan2020} work, which is current state-of-the-art on both LC-QuAD 1.0 and QALD-9 datasets. We also compare our work with GGNN~\cite{webqsp} which is the only known benchmark for WebQSP-WD dataset. We also compare our work with WDAqua~\cite{diefenbach2017wdaqua}, gAnswer~\cite{gAnswer} and QAMP~\cite{qamp} systems. 
\subsection{Results \& Discussion}

Table~\ref{table:results} shows performance of \system ~on two different KBs with five different datasets described in Section \ref{sec:data}. We also show the type of reasoning required for each dataset along with the precision/recall and F1 measure.
We use GERBIL
~\cite{DBLP:journals/semweb/UsbeckRHCHNDU19} to compute performance metrics from the pairs of gold answers and system generated answers from the pipeline.
To our knowledge, ours is the first system reporting KBQA numbers across two different KBs and varying reasoning types. We get state of the art numbers beating NSQA~\cite{pavan2020} on LC-QuAD 1.0 and also achieve comparable numbers for QALD. For WebQSP-WD dataset we get state of the art numbers nearly $20\%$ gain over GGNN~\cite{webqsp}.
The accuracy numbers for Wikidata datasets show that there is ample scope for improvement for different modules. This gets evident when we look at the module performances on a small set of development sets across datasets. For WebQSP-WD we did not have any ground truth for evaluating entity linking and relation linking. 


\noindent\textbf{\textit{AMR:}}
Table \ref{tab:amr_per} show the performance of the AMR parser on the 5 development sets in Smatch (standard measure used to measure AMRs) and Exact Match ($\%$ of questions that match fully with ground truth AMR).

\begin{table}[h]
\begin{small}
\centering
\begin{tabular}{l|c|c|c} 
\toprule
Dataset & KB & Smatch & Exact Match \\\midrule
LC-QuAD 1.0 & DBpedia & 87.6 & 30.0 \\
QALD-9 &   DBpedia & 89.3 & 41.8 \\
WebQSP-WD & Wikidata & 88.0 & 43.8 \\
SWQ-WD & Wikidata & 83.0 & 37.8 \\
TempQA-WD & Wikidata & 89.6 & 39.8 \\
\bottomrule
\end{tabular}
\caption{AMR parser performances on the development sets}
\label{tab:amr_per}
\end{small}
\end{table}

\noindent\textbf{\textit{Entity Linking:}}
Table \ref{EL_performance} below captures the performance of entity linking module on different datasets. The question level accuracy is computed by considering all the entities present in each question. However, mention level accuracy is computed by considering single mention at a time. We can see that entity linking performance for Wikidata datasets is low compared to DBpedia. One of the reasons for is all these datasets are adopted from Freebase dataset and the way entities represented in these two KBs is different. Also, very little research is done on entity linking for Wikidata and KBQA in general. 
\begin{table}[htb]
\begin{small}
\centering
\setlength\tabcolsep{6pt}
\renewcommand{\arraystretch}{1.0}
\begin{tabular}{lcrr}
\toprule
\multirow{2}*{Dataset}& \multirow{2}*{KB} & \multicolumn{2}{c}{Accuracy (\%)}  \\
& & Mention Level & Question Level \\
\midrule
LC-QuAD 1.0 & DBpedia & $86.81\;(91.94)$ &  $84.00\;(90.50)$\\
QALD-9 & DBpedia & $89.52\;(94.28)$ &  $89.79\;(93.87)$\\
TempQA-WD& Wikidata & $74.01\;(82.47)$  & $57.14\;(69.71)$ \\
SWQ & Wikidata & $72.27\;(83.18)$ &  $70.14\;(81.59)$\\
\bottomrule
\end{tabular}
\caption{Entity linking performance on development sets when gold mentions are provided. The numbers inside parentheses denote the Hits@5 scores.}
\label{EL_performance}
\end{small}
\end{table}

\noindent\textbf{\textit{Relation Linking:}} Table \ref{RL_performance} captures the performance of the relation linking module. It demonstrates that relation linking is still a challenging tasks specially with multi-hop reasoning and temporal reasoning, where the query graph is disconnected across events, there is room for improvement.    

\begin{table}[htb]
\begin{small}
\centering
\setlength\tabcolsep{6pt}
\renewcommand{\arraystretch}{1.0}
\begin{tabular}{lcccc}
\toprule
Dataset & KB  & Precision & Recall & F1 \\
\midrule
LC-QuAD 1.0 & DBpedia & 0.52 & 0.50 & 0.50 \\
QALD-9 & DBpedia & 0.55 & 0.53 & 0.53 \\
TempQA-WD& Wikidata & 0.43 & 0.43 & 0.42 \\
SWQ & Wikidata & 0.67 & 0.68 & 0.67 \\
\bottomrule
\end{tabular}
\caption{Relation linking performance on development sets.}
\label{RL_performance}
\end{small}
\end{table}

\subsubsection{Ablation Study:}
To gain more insights on the performance, we also did an ablation study of \system~using \dataset~dev set for Wikidata and 100 question dev set from LC-QuAD 1.0 that we manually annotated with ground truth for all the modules. This shows the impact of individual module on overall performance is evaluated. Table~\ref{table:ablation} shows the results.
For example GT-$AMR$ refers to the case where ground truth AMR is fed directly into $\lambda$-module. The table shows large jump in accuracy (in both the datasets) when fed with ground truth entities (GT-EL) and ground truth relations (GT-RL). This points to the need for improved entity linking and relation linking on both datasets across KBs.

\begin{table}[htb]
\small
\begin{tabular}{l|ccc|ccc}
\toprule
& \multicolumn{3}{|c|}{TempQA-WD} & \multicolumn{3}{|c}{LC-QuAD 1.0} \\
\midrule
 & P & R & F1 & P & R & F1 \\
\midrule
NO GT & 0.47 & 0.50 & 0.47 & 0.50 & 0.52 & 0.50 \\
GT-AMR & 0.50 & 0.51 & 0.50 & 0.51 & 0.53 & 0.51 \\
GT-$\lambda$ & 0.52 & 0.53 & 0.52 & 0.52 & 0.53 & 0.52   \\
GT-EL & 0.60 & 0.62 & 0.60 & 0.63 & 0.66 & 0.64 \\
GT-RL & 0.92 & 0.93 & 0.92 & 0.99 & 0.98 & 0.98 \\
GT-KB-$\lambda$ & 0.93 & 0.93 & 0.93 & 1.0 & 0.99 & 0.99 \\
GT-SPARQL & 1.0 & 1.0 & 1.0 & 1.0 & 1.0 & 1.0 \\
\bottomrule
\end{tabular}
\caption{Ablation Study on \dataset~and LC-QuAD 1.0 dev sets. P-Precision, R-Recall}
\label{table:ablation}
\end{table}

\section{Conclusion}

In this paper, we present \system~system that is generalizable across knowledge bases and reasoning types. We introduce KB-agnostic question understanding component that is common across KBs with AMR based intermediate $\lambda$-Calculus representation. Question Mapping and reasoning module on the specific KB is customized per KB. we also presented rule modules that can aid in adding new reasoning type. 
We introduced a new benchmark dataset \dataset~for temporal KBQA on \textit{Wikidata}. 
Experimental results show that \system ~indeed achieves its generalization goals with state of the art results on LC-QuAD 1.0 and WebQSP-WD.

\bibliography{custom}

\end{document}


\maketitle

\newcommand\tab[1][0.4cm]{\hspace*{#1}}

\subsection{AMR to Lambda Translation}
\begin{table*} \setlength\tabcolsep{7pt}
    \centering
    \begin{small}
    \resizebox{\textwidth}{!}{
    \begin{tabular}{@{}l{} @{~~~} l{} @{~~~} l{}@{}} 
    
        \toprule
        \begin{tabular}[l]{@{}l@{}}
           \textbf{Rule}\\
           \textbf{Type}
        \end{tabular} 
        &
        \textbf{AMR A = (v/frame …)}
        &
        \textbf{Lambda Expression L = $\psi$(v)} 
        \\ 
        \midrule

        \begin{tabular}[l]{@{}l@{}}
           Base
        \end{tabular} 
         &
         \begin{tabular}[l]{@{}l@{}}
            (\textbf{v}/frame :arg0(\textbf{v0}/frame0) … :argn(\textbf{vn}/framen))
        \end{tabular} 
         &
         \begin{tabular}[c]{@{}l@{}}
            frame(\textbf{v}, \textbf{v0}, … \textbf{vn}) $\land$ $\psi$(\textbf{v0}) $\land$ … $\psi$(\textbf{vn}) 
        \end{tabular} 
        \\
         \noalign{\smallskip}
         
         \begin{tabular}[l]{@{}l@{}}
           Base
        \end{tabular} 
         &
         \begin{tabular}[l]{@{}l@{}}
            (\textbf{v}/frame :\textbf{arg1(a/amr-unknown)}  … :argn(vn/framen))
        \end{tabular} 
         &
         \begin{tabular}[c]{@{}l@{}}
           $\lambda$a. $\psi$(\textbf{v})
        \end{tabular} 
        \\
         \noalign{\smallskip}
         
         \begin{tabular}[l]{@{}l@{}}
           Numerical
        \end{tabular} 
         &
         \begin{tabular}[l]{@{}l@{}}
            (\textbf{v}/frame :arg0(\textbf{v0}/frame0 :\textbf{quant(a/amr-unknown})) …\\ :argn(vn/framen))
        \end{tabular} 
         &
         \begin{tabular}[c]{@{}l@{}}
            count($\lambda$\textbf{v0}. $\psi$(\textbf{v}))
        \end{tabular} 
        \\
         \noalign{\smallskip}

         \begin{tabular}[l]{@{}l@{}}
           Numerical
        \end{tabular} 
         &
         \begin{tabular}[l]{@{}l@{}}
           (\textbf{v}/frame :arg0(a/amr-unknown) ... :argn(vn/framen)\\ \textbf{:mod(f/first}))
        \end{tabular} 
         &
         \begin{tabular}[c]{@{}l@{}}
            min($\lambda$a. $\psi$(\textbf{v}), 0, 1)
        \end{tabular} 
        \\
         \noalign{\smallskip}
         
        \begin{tabular}[l]{@{}l@{}}
           Numerical
        \end{tabular} 
         &
         \begin{tabular}[l]{@{}l@{}}
           (\textbf{v}/frame :arg0(a/amr-unknown) ... :argn(vn/framen)\\ \textbf{:mod(f/last}))
        \end{tabular} 
         &
         \begin{tabular}[c]{@{}l@{}}
            max($\lambda$a. $\psi$(\textbf{v}), 0, 1)
        \end{tabular} 
        \\
         \noalign{\smallskip}
         
        \begin{tabular}[l]{@{}l@{}}
          Numerical
        \end{tabular} 
         &
         \begin{tabular}[l]{@{}l@{}}
            (\textbf{v}/frame :arg0(\textbf{v0}/frame0:mod(a/amr-unknown)) …:argn\\
            (\textbf{vn}/frame :arg1-of(h2/\textbf{have-quant-91} :arg3(l/\textbf{most}))))

        \end{tabular} 
         &
         \begin{tabular}[c]{@{}l@{}}
             argmax($\lambda$\textbf{v0}. $\psi$(\textbf{v}), $\lambda$\textbf{v0}. $\lambda$\textbf{vn}. $\psi$(\textbf{vn}), 0, 1)
        \end{tabular} 
        \\
         \noalign{\smallskip}
         
          \begin{tabular}[l]{@{}l@{}}
          Numerical
        \end{tabular} 
         &
         \begin{tabular}[l]{@{}l@{}}
            (\textbf{v}/frame :arg0(\textbf{v0}/frame0:mod(a/amr-unknown)) …:argn\\
            (\textbf{vn}/frame :arg1-of(h2/\textbf{have-quant-91} :arg3(l/\textbf{least}))))

        \end{tabular} 
         &
         \begin{tabular}[c]{@{}l@{}}
             argmin($\lambda$\textbf{v0}. $\psi$(\textbf{v}), $\lambda$\textbf{v0}. $\lambda$\textbf{vn}. $\psi$(\textbf{vn}), 0, 1)
        \end{tabular} 
        \\
         \noalign{\smallskip}
        
         \begin{tabular}[l]{@{}l@{}}
          Numerical
        \end{tabular} 
         &
         \begin{tabular}[l]{@{}l@{}}
            (\textbf{v}/frame :arg0(a/amr-unknown) … :argn(\textbf{vn}/framen :arg1-of\\
            (h2/\textbf{have-degree-91} :arg3(m/\textbf{more}) :arg4(\textbf{vm}/framem\\
            :arg1-of(\textbf{n}/nested-frame))))
        \end{tabular} 
         &
         \begin{tabular}[c]{@{}l@{}}
             $\lambda$a. $\psi$(\textbf{v}) $\land$ $\psi$(\textbf{n}) $\land$ cmp(\textbf{vn}, \textbf{vm}, $>$) 
        \end{tabular} 
        \\
         \noalign{\smallskip}
         
          \begin{tabular}[l]{@{}l@{}}
          Numerical
        \end{tabular} 
         &
         \begin{tabular}[l]{@{}l@{}}
            (\textbf{v}/frame :arg0(a/amr-unknown) … :argn(\textbf{vn}/framen :arg1-of\\
            (h2/\textbf{have-degree-91} :arg3(m/\textbf{less}) :arg4(\textbf{vm}/framem\\
            :arg1-of(\textbf{\textbf{n}}/nested-frame))))
        \end{tabular} 
         &
         \begin{tabular}[c]{@{}l@{}}
             $\lambda$a. $\psi$(\textbf{v}) $\land$ $\psi$(\textbf{n}) $\land$ cmp(\textbf{vn}, \textbf{vm}, $<$) 
        \end{tabular} 
        \\
         \noalign{\smallskip}

         \begin{tabular}[l]{@{}l@{}}
          Temporal
        \end{tabular} 
         &
         \begin{tabular}[l]{@{}l@{}}
            (\textbf{v}/frame :arg0(v0/frame0) … :argn(vn/framen)\\
            \textbf{:time(a/amr-unknown))}
        \end{tabular} 
         &
         \begin{tabular}[c]{@{}l@{}}
          $\lambda$ev. $\psi$(\textbf{v}) $\land$ interval(ev, \textbf{v}) 
        \end{tabular} 
        \\
         \noalign{\smallskip}

        \begin{tabular}[l]{@{}l@{}}
           Temporal
        \end{tabular} 
         &
         \begin{tabular}[l]{@{}l@{}}
            (\textbf{v}/frame :arg0(a/amr-unknown) … :argn(vn/framen)\\
            \textbf{:time(b/before} :op1(\textbf{n}/nested-frame)))
        \end{tabular} 
         &
         \begin{tabular}[c]{@{}l@{}}
           argmax($\lambda$a. $\psi$(\textbf{v}) , $\lambda$a. $\lambda$ev. $\psi$(\textbf{n}) $\land$ interval(ev, \textbf{v}) \\
           $\land$ interval(en, \textbf{n}) $\land$ \textbf{before}(ev, en), 0, 1)
        \end{tabular} 
        \\
         \noalign{\smallskip}

         \begin{tabular}[l]{@{}l@{}}
          Temporal
        \end{tabular} 
         &
         \begin{tabular}[l]{@{}l@{}}
            (\textbf{v}/frame :arg0(a/amr-unknown) … :argn(vn/framen)\\
            \textbf{:time(a/after} :op1(\textbf{n}/nested-frame)))
        \end{tabular} 
         &
         \begin{tabular}[c]{@{}l@{}}
          argmin($\lambda$a. $\psi$(\textbf{v}) , $\lambda$a. $\lambda$ev. $\psi$(\textbf{n}) $\land$ interval(ev, \textbf{v}) \\
          $\land$ interval(en, \textbf{n}) $\land$ \textbf{after}(ev, en), 0, 1)
        \end{tabular} 
        \\
         \noalign{\smallskip}

        \begin{tabular}[l]{@{}l@{}}
           Temporal
        \end{tabular} 
         &
         \begin{tabular}[l]{@{}l@{}}
            (\textbf{v}/frame :arg0(a/amr-unknown) … :argn(vn/framen)\\
            \textbf{:time}(\textbf{n}/nested-frame))
        \end{tabular} 
         &
         \begin{tabular}[c]{@{}l@{}}
           $\lambda$a. $\psi$(\textbf{v}) $\land$ $\psi$(\textbf{n}) $\land$ interval(ev, \textbf{v}) $\land$ interval(en, \textbf{n})\\
           $\land$ \textbf{overlap}(ev, en)
        \end{tabular} 
        \\
         \noalign{\smallskip}

         \begin{tabular}[l]{@{}l@{}}
           Temporal
        \end{tabular} 
         &
         \begin{tabular}[l]{@{}l@{}}
           (\textbf{v}/frame :arg0(a/amr-unknown) :argn(vn/framen)\\ :\textbf{ord(o/ordinal-entity} :value \textbf{x}))
        \end{tabular} 
         &
         \begin{tabular}[c]{@{}l@{}}
          argmin($\lambda$a. $\psi$(\textbf{v}) , $\lambda$a. $\lambda$ev. interval(ev, \textbf{v}), \textbf{x}+1, 1)
        \end{tabular} 
        \\
         \noalign{\smallskip}

         \begin{tabular}[l]{@{}l@{}}
           Temporal
        \end{tabular} 
         &
         \begin{tabular}[l]{@{}l@{}}
           (\textbf{v}/frame :arg0(a/amr-unknown) :argn(vn/framen)\\ :\textbf{ord(o/ordinal-entity} :value \textbf{-1}))
        \end{tabular} 
         &
         \begin{tabular}[c]{@{}l@{}}
          argmax($\lambda$a. $\psi$(\textbf{v}) , $\lambda$a. $\lambda$ev. interval(ev, \textbf{v}), 0, 1)
        \end{tabular} 
        \\
         \noalign{\smallskip}
         
        \begin{tabular}[l]{@{}l@{}}
          Temporal
        \end{tabular} 
         &
         \begin{tabular}[l]{@{}l@{}}
            (\textbf{v}/frame :arg0(a/amr-unknown) … :argn(vn/framen)\\
            \textbf{:time(n/now))}
        \end{tabular} 
         &
         \begin{tabular}[c]{@{}l@{}}
          $\lambda$a. $\psi$(\textbf{v}) $\land$ interval(ev, \textbf{v}) $\land$ interval(en, now()) \\
            $\land$ overlap(ev, en)
        \end{tabular} 
        \\
         \noalign{\smallskip}

        \begin{tabular}[l]{@{}l@{}}
          Temporal
        \end{tabular} 
         &
         \begin{tabular}[l]{@{}l@{}}
            (\textbf{v}/frame :arg0(a/amr-unknown) … :argn(vn/framen)\\
            \textbf{:time(d/date-entity} :month mm :day dd :year yyyy))
        \end{tabular} 
         &
         \begin{tabular}[c]{@{}l@{}}
          $\lambda$a. $\psi$(\textbf{v}) $\land$ interval(en, date(“dd-mm-yyyy”)) \\ $\land$ interval(ev, \textbf{v}) $\land$ overlap(ev, en)
        \end{tabular} 
        \\
         \noalign{\smallskip}

         \begin{tabular}[l]{@{}l@{}}
          Temporal
        \end{tabular} 
         &
         \begin{tabular}[l]{@{}l@{}}
            (\textbf{v}/frame :arg0(a/amr-unknown) … :argn(vn/framen)\\
            \textbf{:time(t/teenager} :domain(\textbf{n}/nested-frame)))
        \end{tabular} 
         &
         \begin{tabular}[c]{@{}l@{}}
          $\lambda$a. $\psi$(\textbf{v}) $\land$ interval(ev, \textbf{v}) $\land$ teenager(en, \textbf{n}) \\
          $\land$ overlap(ev, en)
        \end{tabular} 
        \\
         \noalign{\smallskip}
     
         \begin{tabular}[l]{@{}l@{}}
           Spatial
        \end{tabular} 
         &
         \begin{tabular}[l]{@{}l@{}}
            (\textbf{b}/\textbf{be-located-at-91} :arg0(a/amr-unknown),\\ \textbf{:mod(s/south)} :op1(\textbf{n}/nested-frame))
        \end{tabular} 
         &
         \begin{tabular}[c]{@{}l@{}}
           $\lambda$a. $\psi$(\textbf{b})$\land$  $\psi$(\textbf{n}) $\land$ coordinate(cb, \textbf{b}) $\land$ coordinate(cn, \textbf{n})\\
           $\land$ \textbf{south}(cb, cn)
        \end{tabular} 
        \\
         \noalign{\smallskip}

    \bottomrule
    \end{tabular}
        }
    \end{small}
    \caption{AMR to Lambda Translation}
    \label{tab:amrTranslation}
\end{table*}
Table \ref{tab:amrTranslation} shows the translation rules for transforming AMR frames (high level) into corresponding Lambda expressions. Rule type denote where the rule is applied. We have shown four different types of rules used in the ~\system~. More such rules can be added to support additional reasoning types. These rule templates rely on the AMR constructs to derive the required reasoning functions.
\begin{itemize}
\item \textbf{Base Rules:} These rules are general multi-hop rules where we capture the question meaning in terms of conjunction of different predicates coming from the AMR graph. These also include simple projection of variables that question is expecting to  output or boolean(true/false) output in case of boolean questions. Second base rule gives an example of simple projection based on \textit{amr-unknown}. If there are no \textit{amr-unknown} variables then the question becomes boolean question and its return type is decided as true/false.
\item \textbf{Numerical Rules:} These rules capture all the numerical reasoning that is supported by the system. These rules include simple count questions, min/max or argmin/argmax or comparative questions. We use the AMR modifiers or the frames like \textit{have-degree-91} to derive the required numerical reasoning to be performed for these type of questions. For example AMR's \textit{quant} modifier for the unknown variable results in count operator in $\lambda$-expression, which gets translated to SPARQL count question later in the pipeline. Following are example questions for numerical rules:

\begin{center}
\begin{tabular}{l l}
 count: & \begin{tabular}{l}How many languages are spoken in \\Turkmenistan?\end{tabular}\\ 
 min/max: & \begin{tabular}{l}When did Romney first run for \\president?\end{tabular}\\  
 argmin/argmax: & \begin{tabular}{l}What is the highest mountain in \\Italy?\end{tabular}\\ 
 comparative: & \begin{tabular}{l}Is Lake Baikal bigger than the Great \\Bear Lake?\end{tabular}\\  
\end{tabular}
\end{center}

\item \textbf{Temporal Rules:} These are the rules that capture the temporal constraints coming in the question. We rely on AMRs temporal constructs like \textit{time/date-entity} etc. to decide if a question is temporal or not. Depending on additional constraints on the edge like before/after/ordinal we decide on the kind of reasoning to be performed. In the Table~\ref{tab:amrTranslation}, we described all the temporal rules that we encountered while annotating the TempQA-WD dataset. Since we saw teenager being very prominent construct in many question, we explicitly added a rule to support that reasoning type. We also use synonyms like prior/precedes etc. to be treated similar to  before temporal reasoning rule present in the table. These synonyms are captured as part of the KB-Agnostic rule module. Below are different examples of temporal questions:

\begin{center}
\begin{tabular}{l l}
 overlap: & \begin{tabular}{l}Who was the US president during the \\cold war?\end{tabular}\\ 
 before/after: & \begin{tabular}{l}Who was London mayor before Boris \\Johnson?\end{tabular}\\  
 ordinal: & \begin{tabular}{l}Who was the first host of the tonight \\show?\end{tabular}\\
\end{tabular}
\end{center}

\item \textbf{Spatial Rules:} We just added a single rule to show how special kind of spatial reasoning queries like north of/ above a certain region or south of /below a certain region can be extended into our current framework. Please note that current ~\system~ doesn't have any of these templates added.
\end{itemize}

\subsection{KB Specific Lambda to SPARQL translation}

\begin{table*}[htb] \setlength\tabcolsep{7pt}
    \centering
    \begin{small}
    \resizebox{\textwidth}{!}{%
    \begin{tabular}{@{}l{} @{~~~} l{} @{~~~} l{}@{}} 
        \toprule
        \textbf{Type} &\textbf{ Expression/Predicate E} & \textbf{SPARQL S = $\phi$(E)} \\ 
        \midrule
        
        \begin{tabular}[l]{@{}l@{}}
            $\lambda$ abstraction
        \end{tabular} 
         &
         \begin{tabular}[l]{@{}l@{}}
            $\lambda$x.T
        \end{tabular} 
         &
         \begin{tabular}[c]{@{}l@{}}
          SELECT DISTINCT ?x WHERE \{ $\phi$(T) \}
        \end{tabular} 
        \\
         \noalign{\smallskip}

        \begin{tabular}[l]{@{}l@{}}
          Count expression
        \end{tabular} 
         &
        \begin{tabular}[l]{@{}l@{}}
             count($\lambda$x.T)
        \end{tabular} 
         &
         \begin{tabular}[c]{@{}l@{}}
          SELECT (COUNT(?x) AS ?c) WHERE \{ $\phi$(T) 
          \}
        \end{tabular}
        \\
         \noalign{\smallskip}

        \begin{tabular}[l]{@{}l@{}}
          Argmin expression
        \end{tabular} 
         &
        \begin{tabular}[l]{@{}l@{}}
             argmin($\lambda$x.T$_1$, $\lambda$x.$\lambda$y.T$_2$,\\ \textit{O}, \textit{L})
        \end{tabular} 
         &
         \begin{tabular}[c]{@{}l@{}}
          SELECT DISTINCT ?x WHERE \{ $\phi$(T$_1$) $\phi$(T$_2$) \\
          \} ORDER BY (?yStart) LIMIT \textit{L} OFFSET \textit{O}
        \end{tabular}
        \\
         \noalign{\smallskip}

        \begin{tabular}[l]{@{}l@{}}
          Argmax expression
        \end{tabular} 
         &
        \begin{tabular}[l]{@{}l@{}}
             argmax($\lambda$x.T1, $\lambda$x.$\lambda$y. T2, \textit{O}, \textit{L})
        \end{tabular} 
         &
         \begin{tabular}[c]{@{}l@{}}
          SELECT DISTINCT ?x WHERE \{
           $\phi$(T1)  $\phi$(T2) \\
          \} ORDER BY DESC(?yStart) LIMIT \textit{L} OFFSET \textit{O}
        \end{tabular}
        \\\noalign{\smallskip}

        \begin{tabular}[l]{@{}l@{}}
          Min expression
        \end{tabular} 
         &
        \begin{tabular}[l]{@{}l@{}}
             min($\lambda$x.T, \textit{O}, \textit{L})
        \end{tabular} 
         &
         \begin{tabular}[c]{@{}l@{}}
          SELECT DISTINCT ?x WHERE \{ $\phi$(T) \\
          \} ORDER BY (?xStart) LIMIT \textit{L} OFFSET \textit{O}
        \end{tabular}
        \\
         \noalign{\smallskip}

         \begin{tabular}[l]{@{}l@{}}
          Max expression
        \end{tabular} 
         &
        \begin{tabular}[l]{@{}l@{}}
             max($\lambda$x.T, \textit{O}, \textit{L})
        \end{tabular} 
         &
         \begin{tabular}[c]{@{}l@{}}
          SELECT DISTINCT ?x WHERE \{\\
          \tab $\phi$(T) \\
          \} ORDER BY DESC(?xStart) LIMIT \textit{L} OFFSET \textit{O}
        \end{tabular}
        \\\noalign{\smallskip}

        \begin{tabular}[l]{@{}l@{}}
          KB Predicate
        \end{tabular} 
         &
        \begin{tabular}[l]{@{}l@{}}
            IRI$_p$(r, s$|$IRI$_s$, o$|$IRI$_o$)
        \end{tabular} 
         &
         \begin{tabular}[c]{@{}l@{}}
          ?s$|$IRI$_s$ IRI$_p$ ?o$|$IRI$_o$.
        \end{tabular} 
        \\
         \noalign{\smallskip}

        \begin{tabular}[l]{@{}l@{}}
            Interval predicate\\ for reified facts
        \end{tabular} 
         &
        \begin{tabular}[l]{@{}l@{}}
             wdt:PID(e, s$|$wd:QID$_1$, o$|$wd:QID$_2$)\\
             $\land$ interval(e$^i$, e)
        \end{tabular} 
         &
         \begin{tabular}[c]{@{}l@{}}
          ?s$|$wd:QID$_1$ p:PID ?e. ?e ps:PID ?o$|$wd:QID$_2$.\\
          ?e \textit{pq:P580} ?eiStart. ?e \textit{pq:P582} ?eiEnd.
        \end{tabular} 
        \\
         \noalign{\smallskip}

        \begin{tabular}[l]{@{}l@{}}
            Interval predicate \\for non-reified facts
        \end{tabular} 
         &
        \begin{tabular}[l]{@{}l@{}}
             wdt:PID(e, s$|$wd:QID, x)$\land$\\
             interval(e$^i$, x)
        \end{tabular} 
         &
         \begin{tabular}[c]{@{}l@{}}
          ?s$|$wd:QID wdt:PID ?x.\\
          BIND (?x AS ?eiStart) BIND (?x AS ?eiEnd)
        \end{tabular} 
        \\
         \noalign{\smallskip}

        \begin{tabular}[l]{@{}l@{}}
             Now predicate
        \end{tabular} 
         &
        \begin{tabular}[l]{@{}l@{}}
             now(e$^i$)
        \end{tabular} 
         &
         \begin{tabular}[c]{@{}l@{}}
          BIND (now() AS ?eiStart) BIND (now() AS ?eiEnd)
        \end{tabular} 
        \\
         \noalign{\smallskip}


        \begin{tabular}[l]{@{}l@{}}
          Overlap predicate
        \end{tabular} 
         &
        \begin{tabular}[l]{@{}l@{}}
             overlap(e1$^i$ e2$^i$)
        \end{tabular} 
         &
         \begin{tabular}[c]{@{}l@{}}
          FILTER(?e1iStart<=?e2iEnd \&\&
          ?e2iStart<=?e1iEnd)
        \end{tabular} 
        \\
         \noalign{\smallskip}

        \begin{tabular}[l]{@{}l@{}}
            Before predicate
        \end{tabular} 
         &
         \begin{tabular}[l]{@{}l@{}}
             before(e1$^i$, e2$^i$)
        \end{tabular} 
         &
         \begin{tabular}[c]{@{}l@{}}
          FILTER(?e1iEnd<=?e2iStart)
        \end{tabular}
        \\
        \noalign{\smallskip}

        \begin{tabular}[l]{@{}l@{}}
             After predicate
        \end{tabular} 
         &
        \begin{tabular}[l]{@{}l@{}}
             after(e1$^i$, e2$^i$)
        \end{tabular} 
         &
         \begin{tabular}[c]{@{}l@{}}
          FILTER(?e1iStart>=?e2iEnd)
        \end{tabular}
        \\
        \noalign{\smallskip}

    \bottomrule
    \end{tabular}
    }
    \end{small}
    \caption{Translation of KB Specific Lambda Expression}
    \label{tab:kbSpecificTranslation}
\end{table*}

Each KB-specific $\lambda$-expression construct is mapped to an equivalent SPARQL construct, as rules given in Table~\ref{tab:kbSpecificTranslation}.
$\lambda$-expressions in our setup can be broadly grouped into 6 categories: \textit{$\lambda$ abstraction}, \textit{argmin}, \textit{argmax}, \textit{min}, \textit{max} and, \textit{count} expression. Table \ref{tab:kbSpecificTranslation} gives mapping for each of them. \textit{$\lambda$ abstraction} rules take care of simple multi-hop rules and projection scenarios while translating to SPARQL. Next set of rules like \textit{min/max/argmin/argmax/count} rules provide the base numerical reasoning capabilities. These operators can be clubbed with other types of reasoning. For example, temporal before/after utilize argmax/argmin on date attribute to get the desired effect of sorting and picking the desired entity. 

Each of lambda expressions contains one or more terms such that each term $T_i$ is comprised of one or more predicates connected via $\land$ or $\lor$. Translation of different predicates is also present in Table~\ref{tab:kbSpecificTranslation}. Predicates used in lambda expression can be broadly categorized into \textit{KB predicate}, \textit{interval} predicate, and \textit{temporal} predicate. \textit{KB predicate} is directly mapped to triple pattern in SPARQL, whereas the \textit{interval} predicates create an interval consisting of start time and end time for a given event. Table~\ref{tab:kbSpecificTranslation} has wikidata specific rules which can be used for constructing intervals. \textit{Start time(wdt:P580)}, \textit{end time(wdt:P582)} or \textit{point in time(wdt:P585)}, present in case of reified events, are used for creating the interval. For non-reified events other temporal properties associated with the entities in the triple are used for getting the interval. \textit{Teenager} predicate and \textit{now} make use of \textit{date of birth(wdt:P569)} and \textit{current time(now())} respectively for creating the interval. Each of the wikidata specific rules can be mapped to target KB accordingly. Temporal predicates include \textit{overlap}, \textit{before}, and \textit{after} which make use of SPARQL FILTER condition to filter out the intervals that do not fall under given conditions. 


 \subsection{Dataset Details: Categorization}
\label{sec:category}
\begin{table}[|b]
\begin{small}
\centering
\begin{tabular}{l|l}
\toprule
\begin{tabular}[l]{@{}l@{}}
    \textbf{Category}\\
    \textbf{(dev/Test)}
\end{tabular} 
& 
\textbf{Example} \\
\midrule

\begin{tabular}[l]{@{}l@{}}
    Simple\\
    (92/471)
\end{tabular} 
& 

\begin{tabular}[c]{@{}l@{}}

{\bf When was Titanic movie released?}\\
SELECT ?a WHERE \{ \\
    wd:Q44578 wdt:P577 ?a \}\\
    
   \end{tabular} \\
\hline
\begin{tabular}[l]{@{}l@{}}
    Medium\\
    (71/154)
\end{tabular} 
& 

\begin{tabular}[c]{@{}l@{}}

{\bf who was the US president} \\
{\bf during cold war?   }\\
SELECT DISTINCT ?a WHERE {\\
  ?a wdt:P39 wd:Q11696. \tab
  ?a p:P39 ?e1.  \\
  ?e1 ps:P39 wd:Q11696. \\
  ?e1 pq:P580 ?st1. \tab
  ?e1 pq:P582 ?et1.  \\
  wd:Q8683 wdt:P580 ?st2. \\
  wd:Q8683 wdt:P582 ?et2. \\
FILTER (?st1 <= ?et2 \&\& ?st2 <= ?et1)}
 \\

  \end{tabular} \\
\hline
\begin{tabular}[l]{@{}l@{}}
    Complex\\
    (12/39)
\end{tabular} 
& 
\begin{tabular}[c]{@{}l@{}}

{\bf who was president of the us when }\\
{\bf douglas bravo was a teenager? } \\
SELECT DISTINCT ?a WHERE \{\\
  ?a p:P39 ?e. \tab
  ?e ps:P39 wd:Q11696.\\
  ?e pq:P580 ?st1.\tab
  ?e pq:P582 ?et1.\\
  wd:Q4095606 wdt:P569 ?x.\\
  bind ((?x + "P13Y"$^\wedge^\wedge$xsd:duration) as ?st2)\\
  bind ((?x + "P19Y"$^\wedge^\wedge$xsd:duration) as ?et2)\\
  FILTER (?st1<=?et2 \&\& ?st2<=?et1) \}\\
\end{tabular} \\

\bottomrule
\end{tabular}
\caption{Examples of Simple, Medium, and Complex temporal reasoning question in the dataset.}
\label{tab:complexity_examples}
\end{small}
\end{table}

In our dataset, \dataset, we also labeled questions with complexity category based on the complexity of the question in terms of temporal and non temporal reasoning required to get the answer, 
as defined below.
\vspace{0.1cm}

\noindent \textbf{1) Simple:} Questions that involve one temporal event and need no temporal reasoning to derive the answer. For example, questions involving simple retrieval of a temporal fact or simple retrieval of other answer types using a temporal fact. 

\noindent \textbf{2) Medium:} Questions that involve two temporal events and need temporal reasoning (such as overlap/before/after) using time intervals of those events. We also include those questions that involve single temporal event but need additional non-temporal reasoning.

\noindent \textbf{3) Complex:} Questions that involve two or more temporal events, need one temporal reasoning and also need an additional temporal or non-temporal reasoning like teenager or spatial or class hierarchy.

Table~\ref{tab:complexity_examples} gives further details of the categorization including the number of dev and test questions for each category together with an example question and the corresponding SPARQL query.
\nobibliography{aaai22}